\documentclass[10pt,twocolumn,letterpaper]{article}

\usepackage[]{cvpr}      

\usepackage{graphicx}
\usepackage{amsmath}
\usepackage{amssymb} 
\usepackage{booktabs}
\usepackage{algorithm}
\usepackage{algorithmic}
\usepackage{amsfonts}       
\usepackage{nicefrac}       
\usepackage{microtype}      
\usepackage{xcolor}         
\usepackage{enumitem}
\usepackage{makecell}
\usepackage{bbding}
\usepackage{pifont}
\usepackage{amsthm}
\usepackage[switch]{lineno}
 \usepackage{multirow}
 \usepackage[normalem]{ulem}
 \useunder{\uline}{\ul}{}

\usepackage{colortbl}
\newtheorem{theorem}{Theorem}

\definecolor{gray}{RGB}{222,222,222}

\newcommand{\fl}{FL\,}

\definecolor{cvprblue}{rgb}{0.21,0.49,0.74}
\definecolor{cvprred}{rgb}{0.901,0.384,0.329}
\usepackage[pagebackref,breaklinks,colorlinks,citecolor=cvprblue]{hyperref}
\usepackage[capitalize]{cleveref}
\crefname{section}{Sec.}{Secs.}
\Crefname{section}{Section}{Sections}
\Crefname{table}{Table}{Tables}
\crefname{table}{Tab.}{Tabs.}

\title{FedRA: A Random Allocation Strategy for Federated Tuning to Unleash the Power of Heterogeneous Clients}

\author{Shangchao Su, \, Bin Li\thanks{Corresponding author}, \, Xiangyang Xue\\
Fudan University, Shanghai, China\\
{\tt\small \{scsu20, libin, xyxue\}@fudan.edu.cn}}

\begin{document}
\maketitle

\begin{abstract}
  With the increasing availability of Foundation Models, federated tuning has garnered attention in the field of federated learning, utilizing data and computation resources from multiple clients to collaboratively fine-tune foundation models. However, in real-world federated scenarios, there often exist a multitude of heterogeneous clients with varying computation and communication resources, rendering them incapable of supporting the entire model fine-tuning process. In response to this challenge, we propose a novel federated tuning algorithm, FedRA. The implementation of FedRA is straightforward and can be seamlessly integrated into any transformer-based model without the need for further modification to the original model. Specifically, in each communication round, FedRA randomly generates an allocation matrix. For resource-constrained clients, it reorganizes a small number of layers from the original model based on the allocation matrix and fine-tunes using adapters. Subsequently, the server aggregates the updated adapter parameters from the clients according to the current allocation matrix into the corresponding layers of the original model. It is worth noting that FedRA also supports scenarios where none of the clients can support the entire global model, which is an impressive advantage. We conduct experiments on two large-scale image datasets, DomainNet and NICO++, under various non-iid settings. The results demonstrate that FedRA outperforms the compared methods significantly. The source code is available at \url{https://github.com/leondada/FedRA}.
\end{abstract}

\section{Introduction}

\begin{figure}[tb]
\centering
\includegraphics[width=1.\linewidth]{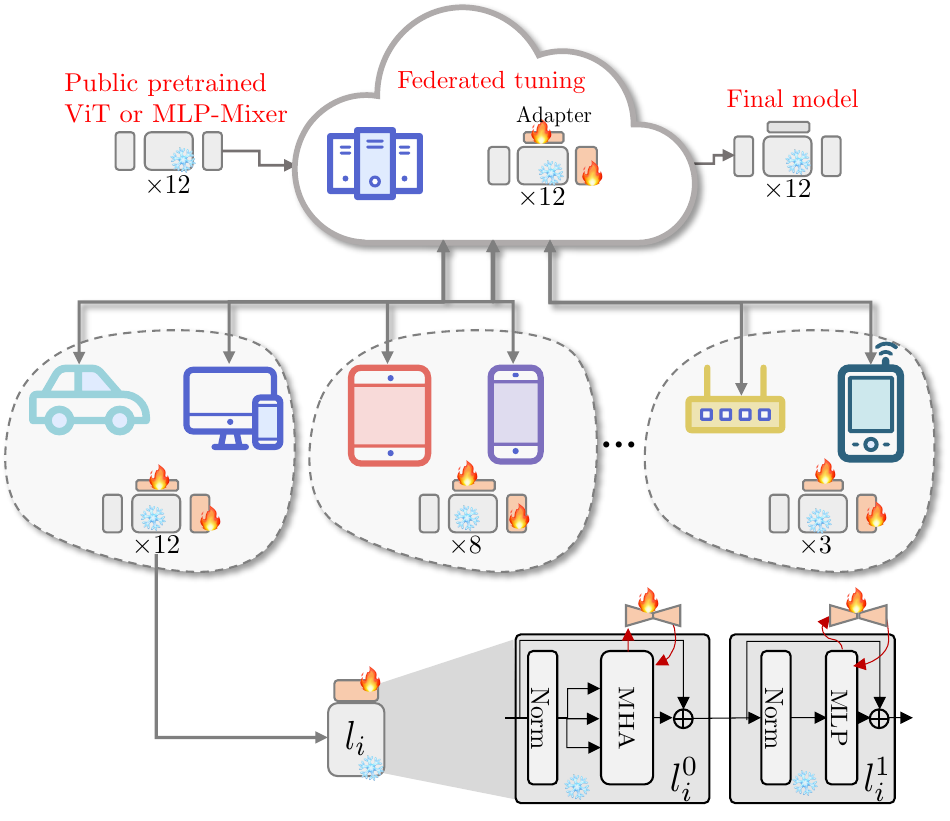}
\caption{Federated tuning for heterogeneous clients.}
\vspace{-10pt}
\label{task}
\end{figure}

Federated learning (\fl)~\cite{mcmahan2017communication}, as a special form of distributed training, allows participants to engage in collaborative learning without sharing private data. A classic framework involves the server sending the initial global model to clients. The clients then train the global model using their local data. The server subsequently performs a weighted average of the updated client model parameters to obtain the new global model for the next round. This classic architecture typically requires training the whole global model parameters for the entire \fl\,procedure.

However, with the increasing popularity of pre-trained foundation models, there are a lot of powerful pre-trained models available, such as ViT~\cite{dosovitskiy2020image}, CLIP~\cite{radford2021learning}, GPT~\cite{brown2020language}, and so on. The classic \fl approach struggles to make effective use of these pre-trained models and thus requires significant communication and computational costs. Therefore, there has been a growing interest in performing federated tuning on top of these foundation models. For instance, some works~\cite{guo2022promptfl,su2022cross} propose prompt fine-tuning algorithms for CLIP, while FedIns~\cite{feng2023towards} and pFedPG~\cite{yang2023efficient} introduce federated tuning algorithms tailored for ViT models. Additionally, FATE-LLM~\cite{fan2023fate} and GPT-FL~\cite{zhang2023gpt} propose using \fl algorithms for fine-tuning large language models.

\begin{figure*}[tb]
\centering
\includegraphics[width=.99\linewidth]{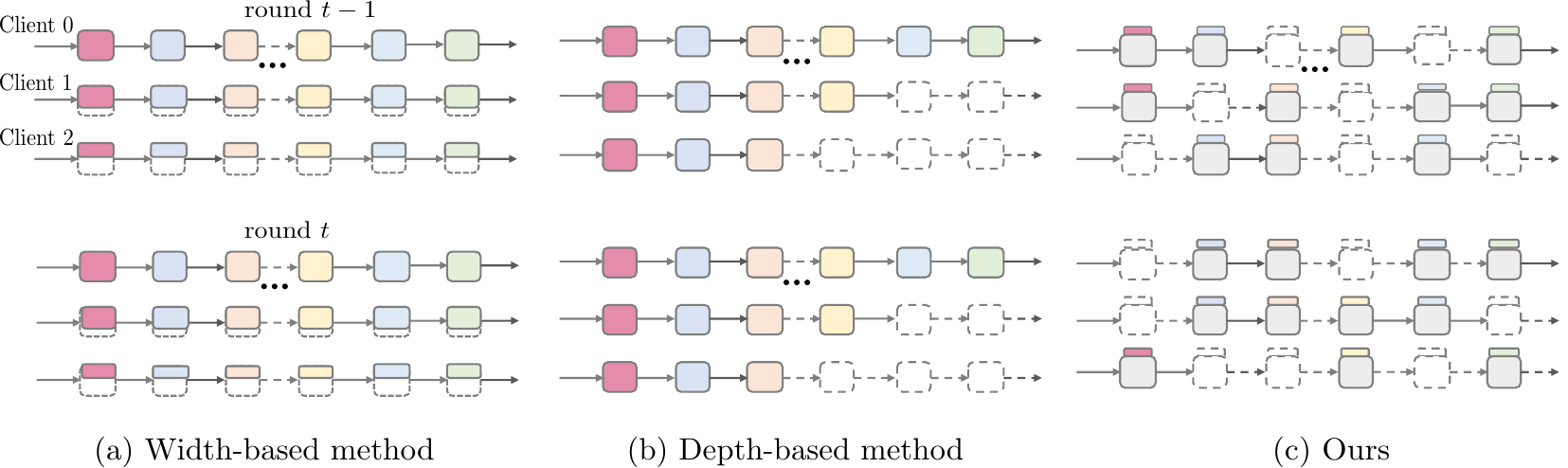}
\caption{(a) The width-based methods disrupt the structure of each layer and therefore cannot be directly applied to pre-trained parameters.  (b) Although depth-based methods can preserve the integrity of the entire model layers, they face a significant issue of feature imbalance, where only a small fraction of resource-rich clients can train the higher layers of the model. (c) Our approach involves the random allocation of adapters, enabling more efficient federated tuning of pre-trained foundation models.}
\vspace{-10pt}
\label{explain} 
\end{figure*}

While federated tuning has improved the efficiency of \fl and, to a certain extent, reduced the computational and communication costs for clients, it is important to note that due to the large parameter size of existing foundation models, many resource-constrained devices are unable to perform fine-tuning on the complete foundation model. Furthermore, in real-world scenarios, such as edge computing~\cite{varghese2016challenges}, clients are heterogeneous, meaning they possess varying computation and communication capabilities, and thus cannot adopt a uniform model structure. In this context, the design of federated tuning algorithms tailored for heterogeneous clients becomes crucial. As shown in Figure~\ref{task}, we introduce the heterogeneous client federated tuning task (HeFT), in this setting, the server needs to interact with a diverse set of clients, leveraging their computational power and data to collaboratively fine-tune the pre-trained foundation model. Compared to the standard heterogeneous client federated learning task, HeFT demands a more nuanced handling of the relationship between the foundation model parameters and federated training.

One approach~\cite{bonawitz2019towards} to handling scenarios involving heterogeneous clients is to discard clients with low computational capabilities and only involve clients with sufficient computational resources in federated tuning. However, this method fails to utilize the data from the discarded clients. The second classical methods~\cite{diao2020heterofl,horvath2021fjord,li2021hermes,ilhan2023scalefl,liao2023adaptive} are based on width pruning, see Figure~\ref{explain} (a), where channels of the global model are pruned to allocate smaller models for low-resource clients. However, such methods forcibly remove a large number of parameters from the pre-training layers, disrupting the mapping relationships of each layer in the foundation model. As a result, they cannot be directly applied to federated tuning. The third approach~\cite{liu2022no,kim2022depthfl}, see Figure~\ref{explain} (b), better suited for federated tuning with heterogeneous clients, employs a technique based on depth pruning. This class of methods allocates only the initial layers of the foundation model to low-resource clients. As they retain complete information from the original model layers, they demonstrate superior performance compared to width-based methods.

However, we observe that in heterogeneous client federated tuning, depth-based methods face a significant challenge. Taking into account the residual connections present in the model, the input features of the model's classification head essentially encompass the outputs of the feature extraction layers at each level of the model. The issue lies in the fact that the features in the shallow layers of the model have already acquired rich information from training on a large number of clients. However, the features in the higher layers of the model have only undergone training on a limited set of resource-rich clients, leading to the classification results of the model being more influenced by clients with limited resources. We term this phenomenon as the feature imbalance problem.

In this context, we propose an exceedingly simple yet effective FedRA algorithm that utilizes {R}andom \textbf{A}llocation to solve the feature imbalance problem. 
Specifically, in each communication round, when the server disseminates the global model, a random allocation strategy is employed, where each client receives random sublayers from the global model (including trainable adapter parameters and initial frozen model parameters) to construct a new local model. After conducting local fine-tuning, the client then transmits the adapter parameters back to the server for aggregation. As depicted in Figure~\ref{explain} (c), random allocation ensures that each layer of the global model learns information from all clients. 
The proposed FedRA algorithm can be applied to all models with repetitive substructures, such as ViT, MLP-Mixer, BERT, etc., and is compatible with all plug-in adapter fine-tuning methods like LoRA. In comparison to existing depth-based methods like InclusiveFL and DepthFL, FedRA is simple to use, requires no additional training loss, and does not necessitate modifications to the model structure.

We conduct experiments using two model architectures, ViT~\cite{dosovitskiy2020image} and MLP-Mixer~\cite{tolstikhin2021mlp}, on two large-scale datasets designed for Non-I.I.D. image classification, DomainNet~\cite{peng2019moment} and NICO++~\cite{he2021towards}. In these experiments, the smallest client model consists of only three layers of feature extraction, while the largest client model encompasses the complete 12-layer architecture. Across various Non-I.I.D. settings, FedRA achieves state-of-the-art performance. Furthermore, FedRA exhibits commendable performance even when no client possesses the entire model. Our contributions are summarized as follows :
\begin{itemize}[noitemsep, leftmargin=*]
\item We introduce the heterogeneous client federated tuning (HeFT) task for the first time. HeFT places higher demands on federated tuning algorithms, requiring them to adapt to the imbalanced computational power of heterogeneous clients while maximizing the utilization of pre-trained knowledge.

\item We propose the novel FedRA algorithm for the HeFT task, which addresses the feature imbalance issue through a random allocation strategy. Even when the computational power of all client devices is insufficient to train a complete global model, FedRA can still work.

\item We conduct experiments on two pre-trained models, ViT and MLP-Mixer, across various large-scale datasets. The results indicate that FedRA outperforms competing methods significantly.

\end{itemize}

\section{Related Work}
\textbf{Federated Learning.}
Federated Learning~\cite{mcmahan2017communication} has rapidly emerged as a popular privacy-preserving collaborative learning algorithm, sparking a significant body of research. Among these, the most prominent is addressing the substantial challenges faced by the FedAvg~\cite{mcmahan2017communication} algorithm under non-iid conditions. Due to significant differences in client data distributions, a considerable amount of information is lost when averaging gradients from different clients on the server, leading to a decline in global model performance. To tackle this issue, FedProx~\cite{li2020federated} and FedDyn~\cite{acar2021federated} propose the incorporation of regularization loss functions during client training to prevent excessive divergence of client models from the global model, thereby mitigating non-iid challenges to some extent. SCAFFOLD~\cite{karimireddy2020scaffold} introduces control variates to correct client drift during local training. In addition to these optimization approaches, some work attempts to address non-iid issues from a personalized perspective. For instance, FedBN~\cite{li2021fedbn} employs local batch normalization to alleviate data disparities among clients, while DFL~\cite{DBLP:conf/icml/LuoWWST22} and FedRep~\cite{collins2021exploiting} decouple the model into personalized and global components for separate learning. Due to the fact that standard FL algorithms require training the model from scratch, they often incur higher communication costs compared to federated tuning.

\textbf{Federated Tuning.}
As the foundation models become increasingly powerful, some efforts have emerged to explore the integration of \fl with these foundation models. This leverages the pre-training knowledge of foundation models to reduce the communication rounds in \fl. Additionally, fine-tuning a few parameters can simultaneously lower the communication costs per round. Approaches such as PromptFL~\cite{guo2022promptfl}, pFedPrompt~\cite{guo2023pfedprompt}, FedCLIP~\cite{lu2023fedclip}, and FedAPT~\cite{su2022cross} have effectively enhanced classification performance while reducing communication costs by designing federated tuning algorithms tailored for CLIP. pFedPT~\cite{li2023visual} and pFedPG~\cite{yang2023efficient} have devised federated prompt tuning algorithms for ViT models, extending the application of foundation models in personalized \fl. FedIns~\cite{feng2023towards} introduces the SSF pool, enabling instance-adaptive inference for \fl. FedPR~\cite{feng2023learning} employs null space learning to mitigate the catastrophic forgetting issue in federated prompt fine-tuning. In natural language tasks, some works~\cite{fan2023fate,zhang2023gpt,lin2023efficient} apply federated tuning to large pre-trained Models. Despite the valuable progress achieved by these works, they have not addressed how federated tuning can be performed on real-world heterogeneous devices.

\textbf{FL for Heterogeneous Clients.}
Existing FL algorithms for heterogeneous clients can be broadly categorized into two types. The first type is width-based approaches~\cite{diao2020heterofl,horvath2021fjord,li2021hermes,ilhan2023scalefl,liao2023adaptive}, which prune channels of the model to allocate lighter models for resource-constrained clients. However, channel pruning disrupts the layer structure of pre-trained models, making it challenging to apply in the federated tuning scenario. The second type is depth-based approaches~\cite{liu2022no,kim2022depthfl}, which retain only the initial layers of the global model for low-resource clients to reduce resource consumption. Among them, InclusiveFL~\cite{liu2022no} compensates for the gradients of lower layers by using the average gradients of higher layers during model aggregation. DepthFL~\cite{kim2022depthfl} requires adding multiple additional classifiers to the global model and trains these classifiers through self-distillation. Finally, the ensemble of multiple classifiers is used as the ultimate classification result. While these methods have achieved certain effectiveness, they have not focused on the new setting of federated tuning and still have room for significant improvement. In contrast, we analyze the shortcomings of existing depth-based methods in the context of federated tuning and propose a simple and efficient FedRA algorithm, which does not require introducing additional model structures or training losses.

\begin{figure}[tb]
\centering
\includegraphics[width=1.0\linewidth]{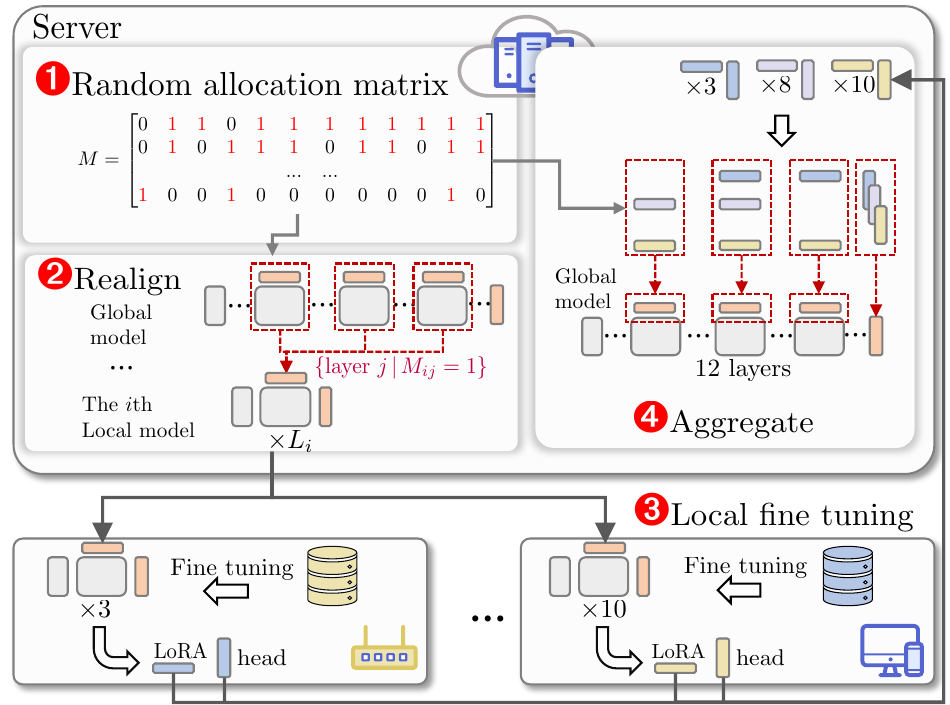}
\caption{The framework of FedRA. In each communication round, the server first randomly generates an allocation matrix to assign subsets of the global model to clients. After client-side fine-tuning, the server collects the fine-tuned LoRA parameters and aggregates them into the global model based on the allocation matrix.}
\vspace{-10pt}
\label{framework}
\end{figure}

\section{The Proposed FedRA}
\subsection{Problem Setting}
Suppose we have $N$ clients each with private data $\mathcal{D}_n$. Assuming the global model has $L$ layers of feature extraction denoted as $\{\boldsymbol{\theta}_i\}_{i=1}^L$. Our objective is to collaboratively leverage the computational and data resources of all clients to perform fine-tuning on the global model. Without loss of generality, we employ the popular LoRA~\cite{hu2021lora} fine-tuning algorithm.
This involves finding the parameters $\{\mathbf{r}_i\}_{i=1}^L$ where $\mathbf{r}_i$ is the LoRA parameters of the $i$-th layer. Due to client heterogeneity, the $n$-th client is only capable of fine-tuning a model with a capacity of $L_n$ layers, denoted as $\{\boldsymbol{\theta}_i, \mathbf{r}_i\}_{i\in\mathcal{S}_n}$, $|\mathcal{S}_n|=L_n$. In the  standard depth-based method, $\mathcal{S}_n= \{1,2,\cdots,L_n\}$.  Then the fine-tuned global model parameters $\{\boldsymbol{\theta}_i, \mathbf{r}_i\}_{i=1}^L$ can minimize the global objective:
\begin{align}
\min _{\{\mathbf{r}_i\}_{i=1}^L} \frac{1}{N} \sum_{n=1}^N \mathbb{E}_{\mathbf{x},y\sim \mathcal{D}_n}[\xi(\mathbf{x}, {y},\{\boldsymbol{\theta}_i, \mathbf{r}_i\}_{i=1}^L)],
\end{align}
where $\xi$ represents the local loss function.

\subsection{The Overall Framework}
The overall FedRA framework is incredibly simple and consists of four steps:

\ding{202} In the first step, we randomly generate an allocation matrix ${\boldsymbol{M} \in \{0,1\}^{N \times L} }$, where $\sum_j \boldsymbol{M} _{ij} = L_i$, indicating that the $i$-th client has $L_i$ layers. At this point, the layers contained in the $i$-th client can be represented as $\mathcal{S}_i =\{j | \boldsymbol{M} _{ij}=1\}$. Note that for each client model, the probability of each layer being selected is $L_i/L$, where $L$ is the total number of layers in global model. Therefore, after $T$ rounds of communication, the expected number of times each layer is selected is $(L_i/L) \times T$.

\ding{203}  In the second step, we utilize the allocation matrix to construct the local models for the current round from the global model, denoted as $\{\boldsymbol{\theta}_j,\boldsymbol{r}_{j}\}_{j\in\mathcal{S}_{i}}$. These models are then dispatched to the clients for training. It is crucial to note that when constructing the local model, the original pre-trained model's layers $\boldsymbol{\theta}_j$ must be dispatched alongside. This step is of paramount importance, if only LoRA parameters are dispatched, the issue of feature imbalance in the global model cannot be addressed.

 \ding{204} The third step involves client fine-tuning, which includes adjusting the LoRA parameters and the classification head of the model. 

 \ding{205} In the fourth step, the server collects the well-trained LoRA parameters from the clients and performs a weighted average of the LoRA according to the allocation matrix: $\boldsymbol{r}_j=\frac{\sum_{i,\boldsymbol{M} _{ij}=1} \boldsymbol{r}_{j}^i|\mathcal{D}_i|}{\sum_i \boldsymbol{M} _{ij}|\mathcal{D}_i|},$
where $\boldsymbol{r}_j^i$ means the $j$-th global model layer collected from the $i$-th client. 
Repeating the above four steps, the expected number of times each layer of the global model is selected is equal.

Notice that in some extreme cases, all clients may be unable to accommodate the complete global model, leading to a possibility of a column in the allocation matrix being entirely filled with zeros. In such scenarios, there are two potential solutions: one involves substituting parameters from the previous round's global model, as:
\begin{align}
\label{eq3}
\boldsymbol{r}_j=\left\{\begin{aligned}\frac{\sum_{i,\boldsymbol{M} _{ij}=1} \boldsymbol{r}_{j}^i|\mathcal{D}_i|}{\sum_i \boldsymbol{M} _{ij}|\mathcal{D}_i|}, \quad& \text{if} \sum_i \boldsymbol{M} _{ij}>0 \\ 
\boldsymbol{r}_j \text{in last round}, \quad& \text{if} \sum_i \boldsymbol{M} _{ij} = 0\end{aligned}\right. ,
\end{align}
the other entails incorporating additional constraints during the generation of the random allocation matrix:
\begin{align}
\label{eq4}
\sum_{i}\boldsymbol{M} _{ij}>0,\quad \sum_{j}\boldsymbol{M} _{ij}= L_i,
\end{align}
that is, when generating an allocation matrix, ensure that each layer of the global model is allocated to at least one client.
As we will demonstrate in the experimental section, these two strategies exhibit similar performance.
  
\subsection{Convergence}

\begin{figure}[tb]
\centering
\includegraphics[width=1.\linewidth]{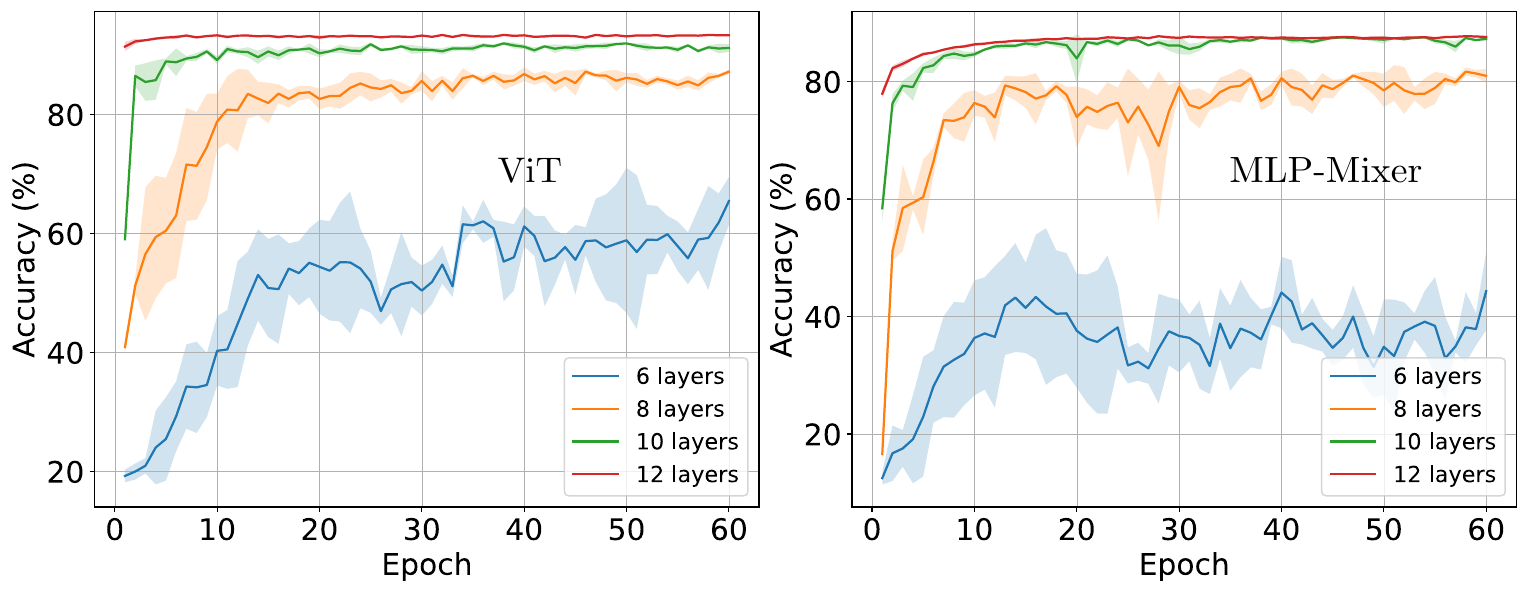}
\caption{Convergence under Random Allocation.}
\vspace{-10pt}
\label{conver}
\end{figure}

We provide experimental and theoretical convergence analyses. First, we design experiments to validate the impact of random allocation on model training. we fine-tune ViT and MLP-Mixer with LoRA on the subset of NICO++~\cite{zhang2023nico++}. In each training epoch, \textbf{we randomly select a subset of the complete model to reconstruct a new model for training. After completing the current epoch, we update the LoRA parameters of these subsets to the corresponding layers of the complete model}. As shown in Figure~\ref{conver}, we can see that even when training only a small portion of the 12 layers in each epoch, the complete model can still converge. Moreover, as the number of subset layers increased, the model's performance also improved.

Then we analyze the convergence of FedRA after multiple rounds of training, and thus analyze why FedRA is superior to standard depth based methods. We assume that the model incorporates residual connections, which is satisfied in the majority of model structures. Under this assumption, we can treat the parameters of the layers excluded from the local model as zero. In other words, all local models are actually global model parameters multiplied by a binary mask. Suppose in the $t$-th round, $\mathbf{r}_t$ denotes the LoRA parameters of the global model, $\mathbf{r}_{t,n,0}=\mathbf{r}_{t}\odot \mathbf{m}_{t,n}$ is the initial local model and $\mathbf{m}_{t,n}$ is the binary mask of the $n$-th local model, $\left\|\mathbf{r}_t-\mathbf{r}_t \odot \mathbf{m}_{t, n}\right\| \leq \alpha\left\|\mathbf{r}_t\right\|^2$. $F$ is the global objective, $F_n$ is the local objective.
Motivated by~\cite{wang2023theoretical}, we use the following three assumptions:
\begin{align}
  \nonumber
&\text{Lipschitz continuity: }\\&\quad\quad\quad \left\|\nabla F_n\left(\mathbf{r}_1\right)-\nabla F_n\left(\mathbf{r}_2\right)\right\| \leq h\left\|\mathbf{r}_1-\mathbf{r}_2\right\| \\
\nonumber
&\text{Bounded variance: } \\&\quad\quad\quad\mathbb{E}\left\|\nabla F_n\left(\mathbf{r}_{t, n, j} ; \mathbf{x}_{n, j}\right)-\nabla F_n\left(\mathbf{r}_{t, n, j}\right)\right\|^2 \leq \sigma^2 \\
\nonumber
&\text{Bounded heterogeneity: }\\&\quad\quad\quad \mathbb{E}\left\|\nabla F_n\left(\mathbf{r}_t\right)-\nabla F\left(\mathbf{r}_t\right)\right\|^2 \leq \delta^2
\end{align}
These assumptions are widely adopted in theoretical analysis in FL.

 \begin{theorem}
 \label{convergence}
 Based on the residual connection assumption, and the above three assumptions. With the client learning rate $\eta$ satisfying $\frac{3 N}{16 J^2 h \Gamma^*}+\frac{N}{6 J h \Gamma^*} \leq \eta \leq \frac{1}{4 J h}$, we have:
 \begin{small}
 \begin{align}
 \nonumber
  \frac{1}{T} &\sum_{t \in[1, T]} \sum_{l \in S^t} \mathbb{E}\left\|\nabla F^l\left(\mathbf{r}_t\right)\right\|^2  \leq\\ &\frac{\mathbb{E}\left[F\left(\mathbf{r}_1\right)\right]}{T \Delta_1}+\frac{h N \alpha}{T \Delta_1 \Gamma^*} \sum_{t \in[1, T]} \mathbb{E}\left\|\mathbf{r}_t\right\|^2 \\
  & +\frac{17 N}{64 J h \Delta_1 \Gamma^*} \sigma^2+\left(\frac{1}{3}+\frac{3}{32 J}\right) \frac{N}{h \Delta_1 \Gamma^*} \delta^2
\label{r:conv}
\end{align}
\end{small}
where $J$ is the client update steps, $S^t$ is the global model layers trained in this round, $T$ is the number of rounds, $\Delta_1=\frac{J \eta}{2}-\frac{3 N}{32 J h \Gamma^*}-\frac{N}{12 h \Gamma^*}$, $\Gamma^*=\min _{t, l} \Gamma_t^l, l \in S^t, \forall t$, $\Gamma_t^l$ represents the number of clients allocated the parameters of layer $l$ in the $t$-th round.
\end{theorem}

The proof can be found in the supplementary materials. From Theorem~\ref{convergence}, we can see: 1) As the round T increase, the upper bound gradually decreases, and ultimately, it is closely tied to $\Gamma^*$. 2) We can explain why FedRA outperforms other methods. In DepthFL and InclusiveFL, each client uses a fixed set of layers, leading to $\Gamma^*$ depending only on the top layer of the model. Due to the fact that the top layer is only trained by a small number of resource-rich clients,$\Gamma^*$ is typically a smaller value. In contrast, in FedRA, due to random allocation, even the top layer of the model involves multiple clients in training. Therefore, the increased $\Gamma^*$ decreases the upper bound of Eq~\ref{r:conv}.

\begin{table*}[]
\center
\resizebox{0.93\textwidth}{!}{%
\begin{tabular}{cclccccccc}
\Xhline{1pt}
\rowcolor{gray}\multicolumn{10}{c}{Model size:  Server$\leftarrow$12, Clients$\leftarrow${[}12,10,8,6,4,3{]}}                                                                                                                                                                                                                                                                                                                                                                                                                                                                                  \\ \Xhline{1pt}
\rowcolor{gray}\multirow{11}{*}{DomainNet} &                            &                     & \begin{tabular}[c]{@{}c@{}}\textit{clipart}\\ (12 layers)\end{tabular} & \begin{tabular}[c]{@{}c@{}}\textit{infograph}\\ (10 layers)\end{tabular} & \begin{tabular}[c]{@{}c@{}}\textit{painting}\\ (8 layers)\end{tabular} & \begin{tabular}[c]{@{}c@{}}\textit{quickdraw}\\  (6 layers)\end{tabular} & \begin{tabular}[c]{@{}c@{}}\textit{real }\\ (4 layers)\end{tabular}  & \begin{tabular}[c]{@{}c@{}}\textit{sketch}\\  (3 layers)\end{tabular} & Average                \\ \cline{2-10} 
                            & \multirow{5}{*}{ViT}       & \textit{AllLarge (ceiling)}            & \textit{85.01}                                                        & \textit{54.73}                                                          & \textit{80.45}                                                         & \textit{73.69}                                                          & \textit{89.92}                                                      & \textit{79.22}                                                       & \textit{77.17}         \\
                            &                            & AllSmall            & 43.86                                                                 & 18.72                                                                   & 35.14                                                                  & 18.01                                                                   & 50.88                                                               & 27.18                                                                & 32.30                  \\
                            &                            & InclusiveFL \footnotesize{(KDD22)\cite{liu2022no}} & {\ul 81.13}                                                           & 38.81                                                                   & {\ul 61.47}                                                            & {\ul 21.83}                                                             & {\ul 76.38}                                                         & {\ul 56.22}                                                          & {\ul 55.97}            \\
                            &                            & DepthFL \footnotesize{(ICLR23)\cite{kim2022depthfl}}    & 80.65                                                                 & {\ul 40.42}                                                             & 59.21                                                                  & 20.36                                                                   & 73.74                                                               & 55.65                                                                & 55.00                  \\
                           &                            & FedRA (Ours)        & \textbf{82.82 }  \color{cvprblue}  {(+1.69)}                                            & \textbf{55.72 }    \color{cvprblue}   (+15.3)                                              & \textbf{75.14}      \color{cvprblue}    (+13.67)                                          & \textbf{25.41 }           \color{cvprblue}      (+3.58)                                    & \textbf{82.41 }        \color{cvprblue}  (+6.03)                                       & \textbf{61.89}   \color{cvprblue}     (+5.67)                                           & \textbf{63.90}   \color{cvprblue}  (+7.93)\\ \cline{2-10} 
                            & \multirow{5}{*}{MLP-Mixer} & \textit{AllLarge (ceiling)}            & \textit{76.99}                                                        & \textit{44.28}                                                          & \textit{71.20}                                                         & \textit{61.58}                                                          & \textit{83.92}                                                      & \textit{67.44}                                                       & \textit{67.57}         \\
                            &                            & AllSmall            & 35.38                                                                 & 14.48                                                                   & 31.69                                                                  & 18.06                                                                   & 44.25                                                               & 22.92                                                                & 27.80                  \\
                            &                            & InclusiveFL \footnotesize{(KDD22)\cite{liu2022no}} & {\ul 71.58}                                                           & 26.22                                                                   & 50.63                                                                  & 11.60                                                                   & 67.18                                                               & 35.88                                                                & 43.85                  \\
                            &                            & DepthFL\footnotesize{(ICLR23)\cite{kim2022depthfl}}    & 70.11                                                                 & {\ul 33.32}                                                             & {\ul 52.07}                                                            & {\ul 13.45}                                                             & {\ul 67.83}                                                         & {\ul 36.42}                                                          & {\ul 45.53}            \\
                            &                            & FedRA (Ours)        & \textbf{76.32}    \color{cvprblue}    (+4.74)                                            & \textbf{41.56}   \color{cvprblue}   (+8.24)                                                & \textbf{64.29}    \color{cvprblue} (+12.22)                                               & \textbf{16.35}   \color{cvprblue}  (+2.9)                                                  & \textbf{74.35}    \color{cvprblue}  (+6.52)                                            & \textbf{47.37 }   \color{cvprblue}  (+10.95)                                            & \textbf{53.37 }  \color{cvprblue} (+7.84) \\ \Xhline{1pt}
\rowcolor{gray}\multirow{11}{*}{NICO++}    &                            &                     & \begin{tabular}[c]{@{}c@{}}\textit{autumn}\\ (12 layers)\end{tabular}  & \begin{tabular}[c]{@{}c@{}}\textit{dim}\\ (10 layers)\end{tabular}       & \begin{tabular}[c]{@{}c@{}}\textit{grass }\\ (8 layers)\end{tabular}    & \begin{tabular}[c]{@{}c@{}}\textit{outdoor}\\  (6 layers)\end{tabular}   & \begin{tabular}[c]{@{}c@{}}\textit{rock}\\ (4 layers)\end{tabular} & \begin{tabular}[c]{@{}c@{}}\textit{water}\\  (3 layers)\end{tabular}  & Average                \\ \cline{2-10} 
                            & \multirow{5}{*}{ViT}       & \textit{AllLarge (ceiling)}            & \textit{92.91}                                                        & \textit{89.83}                                                          & \textit{93.78}                                                         & \textit{90.53}                                                          & \textit{91.14}                                                      & \textit{90.13}                                                       & \textit{91.39}         \\
                            &                            & AllSmall            & 40.57                                                                 & 32.31                                                                   & 44.24                                                                  & 43.52                                                                   & 40.76                                                               & 37.83                                                                & 39.87                  \\
                            &                            & InclusiveFL \footnotesize{(KDD22)\cite{liu2022no}} & {\ul 90.65}                                                           & 81.70                                                                   & 84.70                                                                  & {\ul 77.05}                                                             & {\ul 80.52}                                                         & {\ul 69.23}                                                          & {\ul 80.64}            \\
                            &                            & DepthFL\footnotesize{(ICLR23)\cite{kim2022depthfl}}     & 90.10                                                                 & {\ul 82.07}                                                             & {\ul 84.97}                                                            & 76.27                                                                   & 78.99                                                               & 66.31                                                                & 79.79                  \\
                            &                            & FedRA (Ours)        & \textbf{91.92}   \color{cvprblue}    (+1.27)                                             & \textbf{88.19}    \color{cvprblue}  (+6.12)                                                & \textbf{90.19}   \color{cvprblue}  (+5.22)                                                & \textbf{84.30 }   \color{cvprblue}  (+7.25)                                                & \textbf{87.17 }    \color{cvprblue}   (+6.65)                                          & \textbf{77.12 }   \color{cvprblue}     (+7.89)                                          & \textbf{86.48 }  \color{cvprblue}  (+5.84)\\ \cline{2-10} 
                            & \multirow{5}{*}{MLP-Mixer} & \textit{AllLarge (ceiling)}            & \textit{86.81}                                                        & \textit{82.07}                                                          & \textit{87.11}                                                         & \textit{82.74}                                                          & \textit{84.55}                                                      & \textit{81.85}                                                       & \textit{84.19}         \\
                            &                            & AllSmall            & 39.42                                                                 & 34.15                                                                   & 42.77                                                                  & 42.57                                                                   & 40.71                                                               & 35.64                                                                & 39.21                  \\
                            &                            & InclusiveFL\footnotesize{(KDD22)\cite{liu2022no}} & 82.57                                                                 & 69.74                                                                   & 73.44                                                                  & 65.91                                                                   & 67.19                                                               & 57.41                                                                & 69.38                  \\
                            &                            & DepthFL \footnotesize{(ICLR23)\cite{kim2022depthfl}}    & {\ul 84.50}                                                           & {\ul 73.26}                                                             & {\ul 76.63}                                                            & {\ul 68.90}                                                             & {\ul 71.03}                                                         & {\ul 59.36}                                                          & {\ul 72.28}            \\
                            &                            & FedRA (Ours)        & \textbf{86.59}    \color{cvprblue}    (+2.09)                                            & \textbf{79.50 }    \color{cvprblue}     (+6.24)                                            & \textbf{83.15 }   \color{cvprblue}    (+6.52)                                             & \textbf{74.39 }          \color{cvprblue}    (+5.49)                                       & \textbf{75.51 }   \color{cvprblue}   (+4.48)                                           & \textbf{66.26 }   \color{cvprblue}     (+6.9)                                           & \textbf{77.57 }   \color{cvprblue} (+5.29)\\ \Xhline{1pt}
\end{tabular}}
\caption{The performance of the global model under the \textbf{feature-skew} setting. In this scenario, each domain has one client, and the model structures are all different. The underscore indicates the best-performing method other than FedRA, with the numbers in parentheses representing the improvement of FedRA relative to the underscored value.}
\label{table1}
\end{table*}


 \section{Experiments}
In this section, we conduct extensive experiments using different model architectures and datasets to validate the outstanding performance of FedRA under two types of non-iid scenarios: feature-skew and feature\&label-skew. Additionally, we show the performance of FedRA in extremely heterogeneous scenarios, where the model depth of all clients is smaller than the global model. Finally, we conduct some ablation experiments to further analyze FedRA.

 \subsection{Experimental Setup}
 \textbf{Models.}
Since a large number of publicly available foundation models are based on the transformer architecture, we used a pre-trained ViT with 12 layers of transformers as the global model. It's worth noting that our approach is not limited to ViT models; any model with repetitive substructures can serve as the global model. Therefore, we also conduct experiments using a pre-trained MLP-Mixer model with 12 layers. We utilize publicly available parameters from Timm~\cite{rw2019timm} that were pre-trained on Imagenet-21k~\cite{deng2009imagenet}, specifically ``vit\_base\_patch16\_224" and ``mixer\_b16\_224". During fine-tuning, we default to inserting LoRA for fine-tuning in the last fully connected layer of each substructure.
 
 \textbf{Datasets and Partition.}
It's worth noting that previous works often utilize small datasets like MNIST and CIFAR10 for heterogeneous client training. However, in this paper, as we need to fine-tune the foundation model, we require real image data. Taking into account the need for diverse data distributions in FL, we employ two large-scale datasets: NICO++~\cite{zhang2023nico++} and DomainNet~\cite{peng2019moment}. NICO++ is specially designed for non-iid image classification. The currently available data comprises approximately 90k images spanning 60 categories, encompassing six styles: \textit{autumn}, \textit{dim}, \textit{grass}, \textit{outdoor}, \textit{rock}, and \textit{water}. DomainNet is an even larger multi-domain dataset, containing around 600k images across 345 categories. It consists of six domains: \textit{clipart}, \textit{infograph}, \textit{painting}, \textit{quickdraw}, \textit{real}, and \textit{sketch}. In our specific usage, to reduce training costs, we only utilize the first 100 categories of DomainNet. 

To construct client datasets, we perform two different partitions on the complete dataset:\textbf{ (1) Feature-skew}: We create six clients using data from six domains, with each client having the same class distribution but different image styles. In this setting, the model depths for the six clients were 12, 10, 8, 6, 4, and 3, respectively. The largest model corresponds to the complete 12-layer model, while the smallest model has only 3 layers.
\textbf{(2) Feature\&label-skew}: To introduce label-skew on top of feature-skew, we split the data from each domain into five parts according to a Dirichlet distribution. This results in the creation of 30 clients, with differences in both class distribution and image features among them. In this setting, the first five clients generated from the first domain all adopt a 12-layer structure. The five clients from the sixth domain all use a 3-layer structure.

 \textbf{Implementation Details.}
We implement FedRA using PyTorch on one NVIDIA RTX 3090Ti GPU. The code has been included in the supplementary materials. For all experiments, we conduct 100 rounds of federated learning. In each round, we select 6 heterogeneous clients to participate in training. Each client uses the SGD algorithm with a learning rate of 0.01 to train for 1 epoch. We default to using FedAvg~\cite{mcmahan2017communication} as the basic optimization algorithm, and in the supplementary materials, we also demonstrate the results of using FedDyn~\cite{acar2021federated} as the optimization algorithm.

 \textbf{Baselines.} To demonstrate the effectiveness of FedRA, we compare it with several methods: 
 \textbf{1) AllLarge (ceiling)}: This represents an ideal scenario where client resource constraints are not considered, and clients perform fine-tuning directly using the complete global model. It's important to note that doesn't exist in practice.
\textbf{2) AllSmall}: In this scenario, all clients use the same number of layers as the smallest model.
\textbf{3) InclusiveFL}~\cite{liu2022no}: It follows the standard approach depicted in Figure~\ref{explain}(c) and proposes the momentum distillation to compensate for the gradient of shallow models.
 \textbf{4) DepthFL}~\cite{kim2022depthfl}: Building upon InclusiveFL, this method introduces additional classification heads for different blocks of the model. During client training, self-distillation loss is applied to these heads. Since the original design was for ResNet models, manual adjustment of the classifier insertion position is needed when applied to models like ViT. In our implementation, we treat every three layers of ViT as a block and insert an additional classifier after them.

\begin{table*}[]
\center
\resizebox{0.93\textwidth}{!}{%
\begin{tabular}{cclccccccc}
\Xhline{1pt}
\rowcolor{gray}\multicolumn{10}{c}{Model size: Server$\leftarrow$12, Clients$\leftarrow$ {[}12{]}*5+{[}10{]}*5+{[}8{]}*5+{[}6{]}*5+{[}4{]}*5+{[}3{]}*5}                                                                                                                                                                                                                                                                                                                                                                                     \\ \Xhline{1pt}
\rowcolor{gray}\multirow{11}{*}{DomainNet} &                            &                     & \begin{tabular}[c]{@{}c@{}}\textit{clipart}\\ (12 layers)\end{tabular} & \begin{tabular}[c]{@{}c@{}}\textit{infograph}\\ (10 layers)\end{tabular} & \begin{tabular}[c]{@{}c@{}}\textit{painting}\\ (8 layers)\end{tabular} & \begin{tabular}[c]{@{}c@{}}\textit{quickdraw}\\ (6 layers)\end{tabular} & \begin{tabular}[c]{@{}c@{}}\textit{real}\\ (4 layers)\end{tabular} & \begin{tabular}[c]{@{}c@{}}\textit{sketch}\\ (3 layers)\end{tabular} & Average                \\ \cline{2-10} 
                            & \multirow{5}{*}{ViT}       & \textit{AllLarge (ceiling)}            & \textit{84.67}                                                & \textit{56.70}                                                  & \textit{80.58}                                                & \textit{65.61}                                                 & \textit{89.63}                                            & \textit{78.47}                                              & \textit{75.94}         \\
                            &                            & AllSmall            & 23.74                                                         & 13.44                                                           & 22.96                                                         & 5.45                                                           & 30.76                                                     & 14.18                                                       & 18.42                  \\
                            &                            & InclusiveFL\footnotesize{(KDD22)\cite{liu2022no}} & 80.84                                                         & 39.03                                                           & {\ul 64.93}                                                   & 16.36                                                          & 78.30                                                     & 56.30                                                       & 55.96                  \\
                            &                            & DepthFL\footnotesize{(ICLR23)\cite{kim2022depthfl}}    & {\ul 81.08}                                                   & {\ul 40.87}                                                     & 64.75                                                         & {\ul 19.29}                                                    & {\ul 78.57}                                               & {\ul 58.69}                                                 & {\ul 57.21}            \\
                            &                            & FedRA (Ours)        & \textbf{82.98 }  \color{cvprblue}   (+1.9)                                     & \textbf{51.30 }  \color{cvprblue}    (+10.43)                                     & \textbf{72.25}   \color{cvprblue}     (+7.32)                                   & \textbf{20.45}  \color{cvprblue}    (+1.16)                                      & \textbf{82.31 }  \color{cvprblue}  (+3.74)                                  & \textbf{60.35 }    \color{cvprblue}  (+1.66)                                  & \textbf{61.61}  \color{cvprblue}  (+4.4) \\ \cline{2-10} 
                            & \multirow{5}{*}{MLP-Mixer} & \textit{AllLarge (ceiling)}            & \textit{72.49}                                                & \textit{41.39}                                                  & \textit{69.73}                                                & \textit{48.37}                                                 & \textit{82.08}                                            & \textit{63.10}                                              & \textit{62.86}         \\
                            &                            & AllSmall            & 9.50                                                          & 9.64                                                            & 15.62                                                         & 1.09                                                           & 17.11                                                     & 8.26                                                        & 10.20                  \\
                            &                            & InclusiveFL\footnotesize{(KDD22)\cite{liu2022no}}& {\ul 68.02}                                                   & 30.75                                                           & {\ul 54.22}                                                   & {\ul 11.25}                                                    & {\ul 69.17}                                               & {\ul 36.65}                                                 & {\ul 45.01}            \\
                            &                            & DepthFL\footnotesize{(ICLR23)\cite{kim2022depthfl}}     & 66.52                                                         & {\ul 32.31}                                                     & 51.89                                                         & 9.21                                                           & 67.83                                                     & 34.96                                                       & 43.79                  \\
                            &                            & FedRA (Ours)        & \textbf{69.23}    \color{cvprblue}     (+1.21)                                  & \textbf{34.79 }   \color{cvprblue}     (+2.48)                                    & \textbf{59.12}  \color{cvprblue}     (+4.9)                                     & \textbf{11.56 }   \color{cvprblue}    (+0.31)                                   & \textbf{70.30}  \color{cvprblue}    (+1.13)                                 & \textbf{40.99 }     \color{cvprblue}    (+4.34)                               & \textbf{47.66}  \color{cvprblue}  (+2.65)\\ \hline
\rowcolor{gray}\multirow{11}{*}{NICO++}    &                            &                     & \begin{tabular}[c]{@{}c@{}}\textit{autumn}\\ (12 layers)\end{tabular}  & \begin{tabular}[c]{@{}c@{}}\textit{dim}\\ (10 layers)\end{tabular}       & \begin{tabular}[c]{@{}c@{}}\textit{grass}\\ (8 layers)\end{tabular}    & \begin{tabular}[c]{@{}c@{}}\textit{outdoor}\\ (6 layers)\end{tabular}   & \begin{tabular}[c]{@{}c@{}}\textit{rock}\\ (4 layers)\end{tabular} & \begin{tabular}[c]{@{}c@{}}\textit{water}\\ (3 layers)\end{tabular}  & Average                \\ \cline{2-10} 

                            & \multirow{5}{*}{ViT}       & \textit{AllLarge (ceiling)}            & \textit{93.13}                                                & \textit{88.43}                                                  & \textit{92.60}                                                & \textit{89.48}                                                 & \textit{90.92}                                            & \textit{88.83}                                              & \textit{90.56}         \\
                            &                            & AllSmall            & 29.25                                                         & 24.94                                                           & 34.185                                                        & 30.98                                                          & 29.82                                                     & 29.52                                                       & 29.78                  \\
                            &                            & InclusiveFL\footnotesize{(KDD22)\cite{liu2022no}} & {\ul 91.42}                                                   & 83.03                                                           & 86.25                                                         & {\ul 80.97}                                                    & {\ul 83.01}                                               & {\ul 73.50}                                                 & {\ul 83.03}            \\
                            &                            & DepthFL\footnotesize{(ICLR23)\cite{kim2022depthfl}}    & 90.60                                                         & {\ul 83.23}                                                     & {\ul 86.30}                                                   & 79.72                                                          & 82.92                                                     & 72.48                                                       & 82.54                  \\
                            &                            & FedRA (Ours)        & \textbf{91.86 }     \color{cvprblue}   (+0.44)                                  & \textbf{86.27}  \color{cvprblue}       (+3.04)                                    & \textbf{88.44 }     \color{cvprblue}     (+2.14)                                & \textbf{84.33 }    \color{cvprblue}         (+3.36)                              & \textbf{85.54 }    \color{cvprblue}     (+2.53)                             & \textbf{76.20}     \color{cvprblue}      (+2.7)                               & \textbf{85.44 }  \color{cvprblue} (+2.41)\\ \cline{2-10} 
                            
                            & \multirow{5}{*}{MLP-Mixer} & \textit{AllLarge (ceiling)}            & \textit{83.89}                                                & \textit{77.14}                                                  & \textit{85.34}                                                & \textit{80.44}                                                 & \textit{81.47}                                            & \textit{76.85}                                              & \textit{80.86}         \\
                            &                            & AllSmall            & 13.52                                                         & 13.93                                                           & 19.80                                                         & 17.62                                                          & 15.36                                                     & 12.44                                                       & 15.45                  \\
                            &                            & InclusiveFL\footnotesize{(KDD22)\cite{liu2022no}} & {\ul 82.02}                                                   & {\ul 69.09}                                                     & {\ul 77.35}                                                   & {\ul 69.32}                                                    & {\ul 72.12}                                               & 59.57                                                       & {\ul 71.58}            \\
                            &                            & DepthFL \footnotesize{(ICLR23)\cite{kim2022depthfl}}     & 80.98                                                         & 68.86                                                           & 74.99                                                         & 67.65                                                          & 71.80                                                     & {\ul 59.59}                                                 & 70.64                  \\
                            &                            & FedRA (Ours)        & \textbf{84.00 } \color{cvprblue}    (+1.98)                                     & \textbf{74.50 } \color{cvprblue}    (+5.41)                                       & \textbf{80.55} \color{cvprblue}   (+3.2)                                        & \textbf{72.62 }  \color{cvprblue}     (+3.3)                                     & \textbf{73.25 }    \color{cvprblue}  (+1.13)                                & \textbf{63.92 }  \color{cvprblue}   (+4.33)                                   & \textbf{74.80 }  \color{cvprblue} (+3.22)\\ \Xhline{1pt}
\end{tabular}}
\caption{The performance of the global model under the \textbf{feature\&label-skew} setting. In this scenario, \textbf{each domain has 5 clients, totaling 30 clients}, and the model structures vary across different domains. In this setting, there exist simultaneous differences in both feature distribution and label distribution among the clients. }
\label{table2}
\end{table*}

\begin{figure}[tb]
\centering
\includegraphics[width=1.\linewidth]{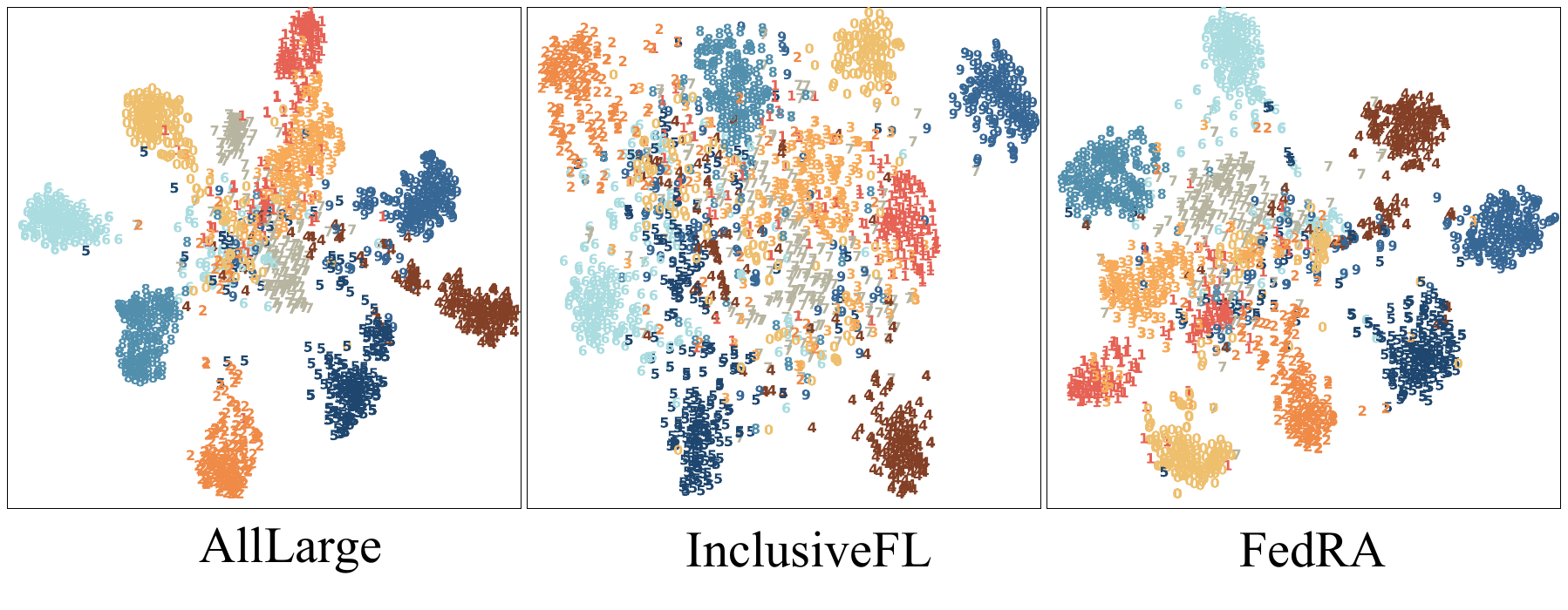}
\caption{The t-SNE visualization of the global model.}
\label{vis}
\end{figure}

\subsection{The Main Results}
\textbf{Feature-skew.}  
Table~\ref{table1} presents the performance of the global model on different domains under the Feature-skew setting. We can see that: 
1) Overall, FedRA's performance is significantly superior to the comparative methods. In DomainNet, whether using ViT or MLP-Mixer, the global model's average accuracy is boosted by 8 points. There is also an improvement of 5 to 6 points in NICO++. 
 2) For clients with different model sizes, FedRA consistently learns knowledge specific to each client. In the testing sets of larger-scale clients (e.g., client0 and client1), FedRA's performance approaches the ceiling. 
3) In the \textit{quickdraw} domain, all methods perform significantly below the ceiling. This is attributed to the substantial stylistic differences in the data of the quickdraw domain compared to other domains, leading to increased learning difficulty.
4) In InclusiveFL and DepthFL, compared to the ceiling AllLarge, there is a noticeable performance drop in models with a layer size of 12. This indicates that these two methods exhibit some disruption with the largest-scale clients when engaged in heterogeneous client collaborative learning. In contrast, FedRA demonstrates minor disruption in such scenarios.
5) We extract some data from 10 classes of DomainNet and visualize the features of the last attention layer in the global model of the ViT using t-SNE~\cite{van2008visualizing}, as shown in Figure~\ref{vis}. FedRA can effectively extract more discriminative features.

\begin{table}[]
\center
\resizebox{\linewidth}{!}{%
\begin{tabular}{clccccccc}
\Xhline{1pt}
 \rowcolor{gray}                          &              & \textit{autumn}                                                   & \textit{dim}                                                      & \textit{grass}                                                    & \textit{outdoor}                                                  & \textit{rock}                                                     & \textit{water}                                                    & Average                                                           \\ \Xhline{1pt}
 \rowcolor{gray}  \multicolumn{9}{c}{Server$\leftarrow$12,Clients$\leftarrow${[}4,4,4,4,4,4{]}}                                                                                                                                                                                                                                                                                                                                                                                                                                                                               \\ \hline
\multirow{2}{*}{ViT}       & InclusiveFL  & 8.08                                                              & 5.64                                                              & 11.48                                                             & 7.19                                                              & 7.64                                                              & 7.99                                                              & 8.00                                                              \\
                           & FedRA (Ours) & \textbf{\begin{tabular}[c]{@{}c@{}}14.18\\ \color{cvprblue}(+6.1)\end{tabular}}   & \textbf{\begin{tabular}[c]{@{}c@{}}10.09\\ \color{cvprblue}(+4.44)\end{tabular}}  & \textbf{\begin{tabular}[c]{@{}c@{}}20.41\\ \color{cvprblue}(+8.93)\end{tabular}}  & \textbf{\begin{tabular}[c]{@{}c@{}}13.51\\ \color{cvprblue}(+6.32)\end{tabular}}  & \textbf{\begin{tabular}[c]{@{}c@{}}12.34\\ \color{cvprblue}(+4.7)\end{tabular}}   & \textbf{\begin{tabular}[c]{@{}c@{}}10.7\\ \color{cvprblue}(+2.71)\end{tabular}}   & \textbf{\begin{tabular}[c]{@{}c@{}}13.54\\ \color{cvprblue}(+5.54)\end{tabular}}  \\ \cline{2-9} 
\multirow{2}{*}{MLP-Mixer} & InclusiveFL  & 6.71                                                              & 6.08                                                              & 6.64                                                              & 5.93                                                              & 5.96                                                              & 7.03                                                              & 6.39                                                              \\
                           & FedRA (Ours) & \textbf{\begin{tabular}[c]{@{}c@{}}23.2\\ \color{cvprblue}(+16.49)\end{tabular}}  & \textbf{\begin{tabular}[c]{@{}c@{}}22.02\\ \color{cvprblue}(+15.93)\end{tabular}} & \textbf{\begin{tabular}[c]{@{}c@{}}26.32\\ \color{cvprblue}(+19.68)\end{tabular}} & \textbf{\begin{tabular}[c]{@{}c@{}}24.51\\ \color{cvprblue}(+18.57)\end{tabular}} & \textbf{\begin{tabular}[c]{@{}c@{}}21.96\\ \color{cvprblue}(+16)\end{tabular}}    & \textbf{\begin{tabular}[c]{@{}c@{}}22.44\\ \color{cvprblue}(+15.41)\end{tabular}} & \textbf{\begin{tabular}[c]{@{}c@{}}23.41\\ \color{cvprblue}(+17.01)\end{tabular}} \\  \Xhline{1pt}
 \rowcolor{gray}  \multicolumn{9}{c}{Server$\leftarrow$12,Clients$\leftarrow${[}6,6,6,6,6,6{]}}                                                                                                                                                                                                                                                                                                                                                                                                                                                                               \\ \hline
\multirow{2}{*}{ViT}       & InclusiveFL  & 21.14                                                             & 17.45                                                             & 22.43                                                             & 21.15                                                             & 21.93                                                             & 19.35                                                             & 20.58                                                             \\
                           & FedRA (Ours) & \textbf{\begin{tabular}[c]{@{}c@{}}72.9\\ \color{cvprblue}(+51.76)\end{tabular}}  & \textbf{\begin{tabular}[c]{@{}c@{}}65.37\\ \color{cvprblue}(+47.92)\end{tabular}} & \textbf{\begin{tabular}[c]{@{}c@{}}74.74\\ \color{cvprblue}(+52.31)\end{tabular}} & \textbf{\begin{tabular}[c]{@{}c@{}}71.6\\ \color{cvprblue}(+50.45)\end{tabular}}  & \textbf{\begin{tabular}[c]{@{}c@{}}73.52\\ \color{cvprblue}(+51.59)\end{tabular}} & \textbf{\begin{tabular}[c]{@{}c@{}}64.18\\ \color{cvprblue}(+44.83)\end{tabular}} & \textbf{\begin{tabular}[c]{@{}c@{}}70.38\\ \color{cvprblue}(+49.81)\end{tabular}} \\ \cline{2-9} 
\multirow{2}{*}{MLP-Mixer} & InclusiveFL  & 19.13                                                             & 15.17                                                             & 16.58                                                             & 17.38                                                             & 18.21                                                             & 18.12                                                             & 17.43                                                             \\
                           & FedRA (Ours) & \textbf{\begin{tabular}[c]{@{}c@{}}57.83\\ \color{cvprblue}(+38.7)\end{tabular}}  & \textbf{\begin{tabular}[c]{@{}c@{}}51.28\\ \color{cvprblue}(+36.11)\end{tabular}} & \textbf{\begin{tabular}[c]{@{}c@{}}58.24\\ \color{cvprblue}(+41.66)\end{tabular}} & \textbf{\begin{tabular}[c]{@{}c@{}}57.28\\ \color{cvprblue}(+39.9)\end{tabular}}  & \textbf{\begin{tabular}[c]{@{}c@{}}55.26\\ \color{cvprblue}(+37.05)\end{tabular}} & \textbf{\begin{tabular}[c]{@{}c@{}}52.3\\ \color{cvprblue}(+34.18)\end{tabular}}  & \textbf{\begin{tabular}[c]{@{}c@{}}55.37\\ \color{cvprblue}(+37.94)\end{tabular}} \\ \Xhline{1pt}
 \rowcolor{gray}  \multicolumn{9}{c}{Server$\leftarrow$12,Clients$\leftarrow${[}10,10,10,10,10,10{]}}                                                                                                                                                                                                                                                                                                                                                                                                                                                                         \\ \hline
\multirow{2}{*}{ViT}       & InclusiveFL  & 78.17                                                             & 75.58                                                             & 81.23                                                             & 77.80                                                             & 77.72                                                             & 76.82                                                             & 77.89                                                             \\
                           & FedRA (Ours) & \textbf{\begin{tabular}[c]{@{}c@{}}91.86\\ \color{cvprblue}(+13.69)\end{tabular}} & \textbf{\begin{tabular}[c]{@{}c@{}}88.75\\ \color{cvprblue}(+13.17)\end{tabular}} & \textbf{\begin{tabular}[c]{@{}c@{}}92.84\\ \color{cvprblue}(+11.61)\end{tabular}} & \textbf{\begin{tabular}[c]{@{}c@{}}91.01\\ \color{cvprblue}(+13.21)\end{tabular}} & \textbf{\begin{tabular}[c]{@{}c@{}}91.23\\ \color{cvprblue}(+13.51)\end{tabular}} & \textbf{\begin{tabular}[c]{@{}c@{}}87.45\\ \color{cvprblue}(+10.63)\end{tabular}} & \textbf{\begin{tabular}[c]{@{}c@{}}90.53\\ \color{cvprblue}(+12.64)\end{tabular}} \\ \cline{2-9} 
\multirow{2}{*}{MLP-Mixer} & InclusiveFL  & 77.46                                                             & 71.22                                                             & 79.59                                                             & 76.36                                                             & 77.41                                                             & 71.56                                                             & 75.60                                                             \\
                           & FedRA (Ours) & \textbf{\begin{tabular}[c]{@{}c@{}}85.76\\ \color{cvprblue}(+8.3)\end{tabular}}   & \textbf{\begin{tabular}[c]{@{}c@{}}80.82\\ \color{cvprblue}(+9.61)\end{tabular}}  & \textbf{\begin{tabular}[c]{@{}c@{}}86.37\\ \color{cvprblue}(+6.79)\end{tabular}}  & \textbf{\begin{tabular}[c]{@{}c@{}}82.98\\ \color{cvprblue}(+6.62)\end{tabular}}  & \textbf{\begin{tabular}[c]{@{}c@{}}83.05\\ \color{cvprblue}(+5.65)\end{tabular}}  & \textbf{\begin{tabular}[c]{@{}c@{}}79.3\\ \color{cvprblue}(+7.74)\end{tabular}}   & \textbf{\begin{tabular}[c]{@{}c@{}}83.05\\ \color{cvprblue}(+7.45)\end{tabular}}  \\ \Xhline{1pt}
 \rowcolor{gray}  \multicolumn{9}{c}{Server$\leftarrow$12,Clients$\leftarrow${[}10,10,6,6,4,4{]}}                                                                                                                                                                                                                                                                                                                                                                                                                                                                             \\ \hline
\multirow{2}{*}{ViT}       & InclusiveFL  & 76.91                                                             & 74.10                                                             & 70.83                                                             & 67.56                                                             & 70.00                                                             & 60.40                                                             & 69.97                                                             \\
                           & FedRA (Ours) & \textbf{\begin{tabular}[c]{@{}c@{}}91.92\\ \color{cvprblue}(+15.01)\end{tabular}} & \textbf{\begin{tabular}[c]{@{}c@{}}87.07\\ \color{cvprblue}(+12.97)\end{tabular}} & \textbf{\begin{tabular}[c]{@{}c@{}}87.06\\ \color{cvprblue}(+16.23)\end{tabular}} & \textbf{\begin{tabular}[c]{@{}c@{}}83.55\\ \color{cvprblue}(+16)\end{tabular}}    & \textbf{\begin{tabular}[c]{@{}c@{}}85.13\\ \color{cvprblue}(+15.14)\end{tabular}} & \textbf{\begin{tabular}[c]{@{}c@{}}76.26\\ \color{cvprblue}(+15.86)\end{tabular}} & \textbf{\begin{tabular}[c]{@{}c@{}}85.17\\ \color{cvprblue}(+15.2)\end{tabular}}  \\ \cline{2-9} 
\multirow{2}{*}{MLP-Mixer} & InclusiveFL  & 72.90                                                             & 65.57                                                             & 67.51                                                             & 62.67                                                             & 64.66                                                             & 56.29                                                             & 64.93                                                             \\
                           & FedRA (Ours) & \textbf{\begin{tabular}[c]{@{}c@{}}85.16\\ \color{cvprblue}(+12.26)\end{tabular}} & \textbf{\begin{tabular}[c]{@{}c@{}}78.1\\ \color{cvprblue}(+12.53)\end{tabular}}  & \textbf{\begin{tabular}[c]{@{}c@{}}79.22\\ \color{cvprblue}(+11.71)\end{tabular}} & \textbf{\begin{tabular}[c]{@{}c@{}}73.73\\ \color{cvprblue}(+11.05)\end{tabular}} & \textbf{\begin{tabular}[c]{@{}c@{}}73.84\\ \color{cvprblue}(+9.17)\end{tabular}}  & \textbf{\begin{tabular}[c]{@{}c@{}}64.57\\ \color{cvprblue}(+8.28)\end{tabular}}  & \textbf{\begin{tabular}[c]{@{}c@{}}75.77\\ \color{cvprblue}(+10.83)\end{tabular}} \\ \Xhline{1pt}
\end{tabular} }
\caption{All client models are smaller than the server model. The blue value represents the improvement relative to InclusiveFL. }
\label{table3}
\end{table}

\textbf{Feature\&label-skew.}  
For the Feature\&label-skew setting, we set the hyperparameter of the Dirichlet distribution to 0.5. As the hyperparameter increases, the non-iid degree of category distribution among clients decreases. When the hyperparameter is 100, the category distribution among clients becomes nearly identical. Due to space constraints, we have included the experimental results for other hyperparameter values in the supplementary material. It is important to note that compared to the Feature-skew setting, the difficulty of this experiment significantly increases. On one hand, the number of clients increases from 6 to 30, and on the other hand, there are differences not only in feature distribution but also in label distribution among clients. In this experiment, there are six models of different sizes among the 30 clients. Table~\ref{table2} presents the results with a hyperparameter of 0.5, showcasing the performance of the global model on testing sets from six different domains. From the table: 1) In comparison to the Feature-skew experiment in Table~\ref{table1}, there is a slight performance decline for all methods, indicating a significant increase in difficulty with the introduction of label-skew. 2) In this setting, FedRA continues to outperform all competing methods. Compared to competitors, FedRA increases the average accuracy by almost 4 points.

\subsection{Extreme heterogeneous scenarios}
To verify the performance in extreme heterogeneous scenarios, we design two experiments on NICO++ dataset:

\textbf{1) All client models are smaller than the server model.} In this setting, existing depth-based methods face a challenging task because some layers of the global model have never been trained. With a server global model of 12 layers, we show the performance differences of the global model with varying client model sizes. As shown in Table~\ref{table3}, when the client model has 10 layers, FedRA is hardly affected and can efficiently train the global model. In extreme cases where the client models have only 4 or 6 layers, all methods experience a performance decline. Nevertheless, FedRA still outperforms InclusiveFL. We also test a mixed scenario where client sizes range from 4 to 10 layers, and in this case, FedRA still exhibits excellent performance. Additionally, it is interesting to note that in the ViT model, we observe a rapid increase in FedRA's performance (from 13.54\% to 70.38\%) when the client model layers increase from 4 to 6, whereas InclusiveFL's performance only returns to normal values (77.89\%) when the number of layers reaches 10.
In order to address the issue of missing layers encountered in this setting, wherein certain layers of the global model have not undergone any training during a particular round of aggregation, we separately validate the two strategies mentioned in Eq.~\ref{eq3} and Eq.~\ref{eq4}. Table~\ref{table7} indicates that these two strategies exhibit similar performance, with Eq.~\ref{eq3} showing slightly better performance.

\begin{figure}[tb]
\centering
\includegraphics[width=0.95\linewidth]{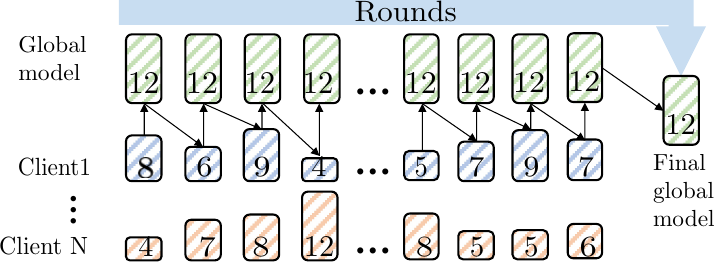}
\caption{The dynamic heterogeneous setting, with each client dynamically changing the model size for each global round.}
\label{heter}
\end{figure}

\textbf{2) The dynamic heterogeneous setting.} To further validate FedRA's performance, we introduce the concept of ``dynamic heterogeneity" as illustrated in Figure~\ref{heter}. In each communication round, the resources available to clients dynamically change, impacting the model structures they can accommodate. To simulate this scenario, we randomly generated the number of layers for client models in each round, \(L_i\in[1,12]\). Table~\ref{table4} presents the results of dynamic heterogeneity experiments, showing that FedRA maintains a leading performance even under dynamic heterogeneity. Additionally, we can observe that, compared to the experiments in Table~\ref{table1} and Table~\ref{table2}, the gap between InclusiveFL and FedRA is reduced in this scenario. This is because, in dynamic structures, even without random allocation, each layer of the global model can access the knowledge from all clients. This further supports the motivation behind FedRA, indicating that maximizing the access of each layer of the global model to client knowledge during fine-tuning leads to a notable enhancement in the overall performance of the global model.

\begin{table}[]
\center
\resizebox{\linewidth}{!}{%
\begin{tabular}{llllllll}
\Xhline{1pt}
 \rowcolor{gray}            & \textit{autumn}                                                   & \textit{dim}                                                      & \textit{grass}                                                 & \textit{outdoor}                                                  & \textit{rock}                                                     & \textit{water}                                                    & Average                                                           \\ \Xhline{1pt}
AllLarge     & \textit{92.91}                                                    & \textit{89.83}                                                    & \textit{93.78}                                                 & \textit{90.53}                                                    & \textit{91.14}                                                    & \textit{90.13}                                                    & \textit{91.39}                                                    \\
AllSmall     & 40.57                                                             & 32.31                                                             & 44.24                                                          & 43.52                                                             & 40.76                                                             & 37.83                                                             & 39.87                                                             \\
InclusiveFL  & 90.82                                                             & \textbf{87.79}                                                             & 92.25                                                          & 86.91                                                             & 89.83                                                             & 84.77                                                             & 88.73                                                             \\
FedRA & \textbf{\begin{tabular}[c]{@{}l@{}}91.86 \\\color{cvprblue} (+1.04)\end{tabular}} & {\begin{tabular}[c]{@{}l@{}}87.75 \\ \color{cvprred}(-0.04)\end{tabular}} & \textbf{\begin{tabular}[c]{@{}l@{}}92.65 \\ \color{cvprblue}(+0.4)\end{tabular}} & \textbf{\begin{tabular}[c]{@{}l@{}}89.48 \\ \color{cvprblue}(+2.57)\end{tabular}} & \textbf{\begin{tabular}[c]{@{}l@{}}91.28 \\ \color{cvprblue}(+1.45)\end{tabular}} & \textbf{\begin{tabular}[c]{@{}l@{}}85.39 \\ \color{cvprblue}(+0.62)\end{tabular}} & \textbf{\begin{tabular}[c]{@{}l@{}}89.74 \\\color{cvprblue} (+1.01)\end{tabular}} \\ \Xhline{1pt}
\end{tabular}}
\caption{The performance of the global model in dynamic Heterogeneous settings. In each communication round, the model layers for each client are randomly selected from 1 to 12. }
\label{table4}
\end{table}

\begin{table}[]
\center
\resizebox{\linewidth}{!}{%
\begin{tabular}{lccccccc}
\Xhline{1pt}
 \rowcolor{gray}           & \textit{autumn} & \textit{dim} & \textit{grass} & \textit{outdoor} & \textit{rock} & \textit{water} & Average \\ \Xhline{1pt}
            & \multicolumn{7}{c}{Server$\leftarrow$12,Clients$\leftarrow${[}4,4,4,4,4,4{]}}                                                       \\ \hline
FedRA(Eq.~\ref{eq3}) & 14.18           & 10.09        & 20.41          & 13.51            & 12.34         & 10.70          & 13.54   \\
FedRA(Eq.~\ref{eq4}) & 13.96           & 9.89         & 17.17          & 13.03            & 12.92         & 10.39          & 12.89   \\ \hline
            & \multicolumn{7}{c}{Server$\leftarrow$12,Clients$\leftarrow${[}6,6,6,6,6,6{]}}                                                       \\ \hline
FedRA(Eq.~\ref{eq3}) & 72.90           & 65.37        & 74.74          & 71.60            & 73.52         & 64.18          & 70.38   \\
FedRA(Eq.~\ref{eq4}) & 71.14           & 61.45        & 72.55          & 68.96            & 71.35         & 60.27          & 67.62   \\ \hline
            & \multicolumn{7}{c}{Server$\leftarrow$12,Clients$\leftarrow${[}10,10,10,10,10,10{]}}                                                 \\ \hline
FedRA(Eq.~\ref{eq3}) & 91.86           & 88.75        & 92.84          & 91.01            & 91.23         & 87.45          & 90.53   \\
FedRA(Eq.~\ref{eq4}) & 92.14           & 89.23        & 92.57          & 90.74            & 91.01         & 87.56          & 90.54   \\ \hline
            & \multicolumn{7}{c}{Server$\leftarrow$12,Clients$\leftarrow${[}10,10,6,6,4,4{]}}                                                     \\
FedRA(Eq.~\ref{eq3}) & 91.92           & 87.07        & 87.06          & 83.55            & 85.13         & 76.26          & 85.17   \\
FedRA(Eq.~\ref{eq4}) & 91.97           & 86.95        & 86.69          & 83.52            & 85.04         & 73.91          & 84.68   \\ \Xhline{1pt}
\end{tabular}}
\caption{Different strategies for handling missing layers. }
\label{table7}
\end{table}

\begin{table}[]
\center
\resizebox{\linewidth}{!}{%
\begin{tabular}{clccccccc}
\Xhline{1pt}
\rowcolor{gray}\multicolumn{9}{c}{Model size: Server$\leftarrow$12,Clients$\leftarrow$ {[}12,10,8,6,4,3{]}}                                                                                                 \\ \Xhline{1pt}
            \rowcolor{gray}                    &             & \textit{autumn} & \textit{dim}   & \textit{grass} & \textit{outdoor} & \textit{rock}  & \textit{water} & Average        \\ \hline
                                
\multirow{2}{*}{$l_i^0 \&l_i^1$} & InclusiveFL & 90.65           & 81.70          & 84.70          & 77.05            & 80.52          & 69.23          & 80.64          \\

                                & FedRA       & \textbf{91.92}  & \textbf{88.19} & \textbf{90.19} & \textbf{84.30}   & \textbf{87.17} & \textbf{77.12} & \textbf{86.48} \\ \hline
                                
\multirow{2}{*}{$l_i^0$}        & InclusiveFL & 90.05           & 81.75          & 85.22          & 79.03            & 82.29          & 72.66          & 81.83          \\

                                & FedRA       & \textbf{91.48}  & \textbf{88.75} & \textbf{89.60} & \textbf{85.02}   & \textbf{86.67} & \textbf{76.78} & \textbf{86.38} \\ \hline
                                
\multirow{2}{*}{$l_i^1$}        & InclusiveFL & 89.99           & 82.07          & 85.66          & 78.04            & 82.15          & 70.11          & 81.34          \\

                                & FedRA       & \textbf{91.15}  & \textbf{87.06} & \textbf{89.30} & \textbf{83.97}   & \textbf{85.22} & \textbf{77.19} & \textbf{85.65} \\ \Xhline{1pt}
\end{tabular}}
\caption{The influence of the insertion position of LoRA. }
\label{table5}
\end{table}

\subsection{More Results}
Here, we analyze the impact of the insertion position and the scale of LoRA on the final results: 
\textbf{1) Insertion Position of LoRA.} In Figure~\ref{explain}(a), we initially assume the insertion of LoRA in both $l_i^0$ and $l_i^1$ of ViT. Here, we explore the effects of inserting LoRA exclusively in $l_i^0$ and $l_i^1$ on different methods. The results are shown in Table~\ref{table5}, indicating that FedRA's performance is robust to the insertion position. Regardless of the position, FedRA consistently outperforms other methods.
\textbf{2) the scale of LoRA.} We also conduct experiments on the scale of LoRA. The results included in the supplementary materials show that all methods maintain stable performance.


\section{Conclusions}
As foundation models become increasingly crucial, federated tuning is becoming increasingly important. In this paper, we introduce FedRA, a federated tuning algorithm tailored for heterogeneous clients. Compared to traditional depth-based heterogeneous \fl algorithms, FedRA is both simple and efficient. Through a random allocation approach, FedRA successfully addresses the issue of feature imbalance across different layers in the global model. Moreover, thanks to this mechanism, FedRA is adaptable to various extreme scenarios of client heterogeneity. We conduct extensive experiments on ViT and MLP-Mixer models using the DomainNet and NICO++ datasets. The results demonstrate that FedRA outperforms comparative methods and exhibits significant potential for application across diverse scenarios.

{
    \small
    \bibliographystyle{ieeenat_fullname}
    \bibliography{main}
}


\end{document}